# Automated simulation and verification of process models discovered by process mining

Ivona Zakarija, Frano Škopljanac-Mačina & Bruno Blašković

Published online: 16 Mar 2020.

Submit your article to this journal 🗗

Article views: 825

View related articles 🗗

View Crossmark data 🗗

Citing articles: 1 View citing articles 🗗

# Automated simulation and verification of process models discovered by process mining

Ivona Zakarija[a], Frano Škopljanac-Mačina[b] and Bruno Blašković[b]

[a]Department of Electrical Engineering and Computing, University of Dubrovnik, Dubrovnik, Croatia; [b]Faculty of Electrical Engineering and Computing, University of Zagreb, Zagreb, Croatia

## ABSTRACT

This paper presents a novel approach for automated analysis of process models discovered using process mining techniques. Process mining explores underlying processes hidden in the event data generated by various devices. Our proposed Inductive machine learning method was used to build business process models based on actual event log data obtained from a hotel's Property Management System (PMS). The PMS can be considered as a Multi Agent System (MAS) because it is integrated with a variety of external systems and IoT devices. Collected event log combines data on guests stay recorded by hotel staff, as well as data streams captured from telephone exchange and other external IoT devices. Next, we performed automated analysis of the discovered process models using formal methods. Spin model checker was used to simulate process model executions and automatically verify the process model. We proposed an algorithm for the automatic transformation of the discovered process model into a verification model. Additionally, we developed a generator of positive and negative examples. In the verification stage, we have also used Linear temporal logic (LTL) to define requested system specifications. We find that the analysis results will be well suited for process model repair.

## 1. Introduction and related work

Process mining is a novel discipline that discovers processes as sequences of events concealed in the vast data logs [1]. Therefore, process mining can be helpful in today's highly connected world where we want to discover or mine behaviour of Multi Agent Systems (MAS) and Internet of Things (IoT) devices. For that purpose, process mining is used to convert vast amounts of event data to a process model. This can be also considered as a *Big Data* [2,3] problem. Process model is a formal or semi-formal representation of underlying processes behaviour, performance and conformance [4]. Authors in [5] coined a term *Internet of Events* or IoE that refers to all event data available online. Therefore, event data are gathered from diverse sources, e.g. from websites and enterprise information systems (Internet of Content), on social networks (Internet of People) or by various interconnected machines and devices (Internet of Things). Data scientists, such as process mining experts, can use discovered process models to explore and better understand, and extract actionable knowledge from the collected event data [5].

This paper will focus on the proposed model checking techniques for automated analysis of the discovered process models. We refer readers to our previous work for additional details on the process model discovery methods [6].

A systematic overview regarding research in the field of process mining is presented in [7], which identified the main research subjects and most relevant algorithms. Furthermore, authors addressed application domains in different business segments. Most of the research papers deal with process discovery, followed by conformance checking (concerning analysis of process models), architecture and tools enhancements. Consequently, it is evident that, relative to process discovery, there are still far fewer techniques and tools for analysis of process models and process repair. The recent approaches dealing with the process repair are described in [8,9].

A methodical review and comparative evaluation of automated process discovery approaches is presented in [10]. The evaluation was conducted using publicly available benchmarks, event log data obtained from information systems and nine quality measures. Problems and unexplored drawbacks in the process mining area are emphasized, including limited scalability of some techniques and large differences in their performance with respect to different quality measures.

A review of the case studies in [11] from different perspectives reveals the state of the implemented process mining applications in the healthcare domain. The application of these techniques is still challenging, especially in healthcare, where processes are inherently

**CONTACT**  Ivona Zakarija  ✉ ivona.zakarija@unidu.hr  Department of Electrical Engineering and Computing, University of Dubrovnik, Dubrovnik, Croatia complex, changeable, dynamic and multi-disciplinary in nature.

Considering that the quality assessment of discovered process models is important for conducting research, as well as for the use of process mining in practice, in [12] a multi-dimensional quality assessment is presented. The results of this study show that evaluation based on real-life event logs considerably distinguishes from the traditional approach to assess process discovery algorithms relied on artificial event logs.

In [13], a scientific framework for supporting the design and execution of process mining workflows is proposed, and generic building blocks required for process mining and various analysis scenarios are described.

Authors in [14] propose an approach for analysing compliance of business processes by utilizing formal methods and verification techniques. Therefore, they introduce specification language, which is formally founded on temporal logic and facilitates the abstract pattern-based specification of compliance requirements.

Methodology for analysing a log containing several traces labelled as compliant or non-compliant is proposed in [15]. It is a logic-based approach that relies on the Inductive Constraint Logic algorithm by utilizing ConDec graphical language for the declarative specification of business processes which adopts an underlying semantics using Linear Temporal Logic (LTL).

In order to verify whether the observed behaviour conforms the expected desirable behaviour authors in [16] developed new language based on LTL.

The approach presented in [17] introduces a new language that checks the conformance of a trace by simply using a Prolog interpreter. A process model is represented as a set of integrity constraints.

Authors in [18] introduced an approach based on temporal logic query checking which stands in the middle between process discovery and conformance checking. The proposed technique produces a set of LTL-based business rules.

A validation methodology for business processes using model-checking techniques is presented in [19]. Business processes are described in a graphical language AMBER with a causality-based semantics.

In this paper, we propose a process mining method for process synthesis, analysis and repair. The proposed method is composed of data preparation, process discovery, analysis and repair of the discovered process model. The main contribution of this paper is our approach for automated analysis of discovered process models based on model checking. An algorithm for automatic transformation of the process model into a verification model is introduced. Additionally, we developed a generator of positive and negative examples. Analysis of the process mining results is performed by simulating process model executions based on real-life event logs and by verification using the Spin model checker. In the verification stage, we have also used LTL to define requested system specifications. Despite the several studies regarding analysis of the process mining results to which we referred above, no one to the best of our knowledge, has provided approach in the manner described in this paper. We differentiate our work by the fact that the above approaches introduce new language for specifications, declarations and analysis. Moreover, as stated in the study [12], the majority of techniques for analysis of discovered process models are based on Petri nets and XML, and furthermore, they are implemented as ProM plug ins. Unfortunately, the ProM tool does not provide dedicated support for the analysis of process mining results given in other notations.

Considering that process mining is a relatively new discipline, there are still many challenges and open questions. The main issues, challenges and problems in process mining are defined in the Manifesto [20]. An overview in [21] provides information on whether defined challenges have been solved or retained open. IEEE Task Force Initiative [22] established competitions with various challenges that promote research and application of process mining [23–25]. In [26], current state and future directions of process mining are reported, also, as one of the problems emphasized the lack of conformance checking techniques.

The rest of the paper is structured as follows: in the second section, we give an overview of our method for process synthesis, analysis and repair, while in the third section, we introduce our use case scenario of a hotel's data management system and describe the collected event log data used for process mining. In the fourth section, we propose and describe in detail our method for automated analysis of discovered process models with the model checking approach. Next, in the fifth section we present and discuss simulation and verification results of the analysis. Finally, in the sixth section we conclude the paper.

## 2. Methodology

In this paper, we propose a novel process mining method for process synthesis, analysis and repair. Process synthesis method is based on Inductive machine learning algorithms and techniques. Model checking is then used to analyse and evaluate results of the process synthesis method. A verification model of the discovered process is built automatically, which can then be simulated and formally checked for specifications conformance using Spin model checker. Process analysis results will be used to refine and repair the discovered process model. We implemented Inductive machine learning algorithm using *libalf* library that includes a variety of learning techniques and algorithms based on finite automata theory [27,28]. For the process discovery, we selected the *k-tail* algorithm, also known as Biermann's algorithm [29], and adapted it for synthesizing processes from behaviour patterns that are recorded in event log data. The *k-tail* algorithm is based on the equivalence relation among states and merging of the equivalent states.

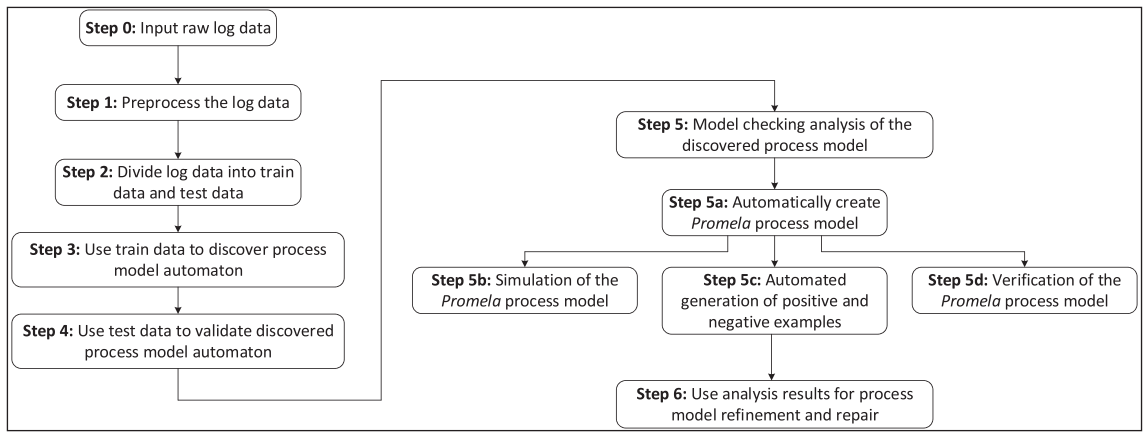

**Figure 1.** Overview of the proposed method for process model synthesis, analysis and repair.

The overview of the proposed method for process synthesis, analysis and repair is presented in Figure 1. Starting from raw event log data, through several automated model transformations, the proposed process synthesis method discovers the process model in labelled transition system notation (Steps 0–4). The subsequent, fifth step of the method with the automated analysis of the discovered process model will be described in detail in the fourth section of this paper.

It must be noted that the initial event data extraction is not trivial, because logs contain only sample behaviour and can be incomplete [6]. Furthermore, the event log data may be too detailed in some instances, and in others it can contain a lot of noise [20,30]. Noise in the event log data refers to incorrectly recorded sequences of events or *traces* in the actual process execution and does not represent typical process. Nevertheless, if the event log noise is omitted, the process discovery algorithm could produce a process model that does not fully conform to actual behaviour recorded in the original event log [31]. The delicate task of the process mining expert is to perform necessary cleaning and filtering of the actual log data (Step 1), while ensuring consistency of the real event logs [32]. Therefore, the process mining expert will include in the training and the test data only representative frequent traces that are most relevant to the actual process.

## 3. Overview of the event log data

As the use case for demonstrating the results of our proposed method, we used in this paper actual log data obtained from the Property Management System (PMS) of a hotel. It is an information system for the management of accommodation facilities, e.g. hotels and resorts, and services organized around the accommodation. The PMS is integrated with a variety of external systems and IoT components [33,34]. Therefore, the PMS can be considered as an MAS system comprised of many interconnected components (IoT devices). Each component acts as an agent within the MAS [35,36]. Most hotels usually offer services such as telephone and Internet access, and contemporary hotels also utilize more advanced IoT systems, e.g. intelligent rooms, Smart TVs or the One Card System.

These are some of the typical PMS–MAS interfaces to IoT systems and devices in contemporary hotels:

- *Intelligent room*: The guest arrival announcement changes heating/cooling mode from standby to work. Guest's RFID card unlocks the room doors and enables power supply, lights and air-conditioning in the room. Maids and technicians can report on the status of the room: e.g. *defect*, *repaired*, *cleaned*.
- *Access control*: RFID bracelet or card signals are detected when the guests move through the hotel, this can be used to charge extra services, e.g. sauna or pool.
- *Wi-Fi Internet*: After guests check in, they receive information about Internet tariffs.
- *Smart TV*: Following check in, it can display a personalized welcome message to the guest and unlock the TV programs. It can also give an overview of expenses, e.g. for pay video content.
- *Telephone exchange*: Receives basic information about conducted telephone calls, (un)locks the extension, and can be used to order automated wake up services.

The central PMS event log contains instances of the guest stay monitoring process, which is a segment of business activities of the hotel reception desk. Data on activities captured from telephone exchange, as well as other external systems and IoT devices, have also been incorporated in collected event log data [37]. Overall, data resources are 15 distinct users who perform guest monitoring and interaction tasks, some of them are hotel employees, while others represent the connected IoT devices.

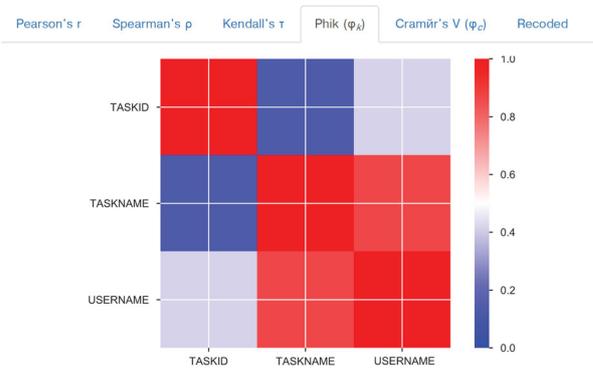

| | DATETIME | TASKID | TASKNAME | USERNAME |
|---|---|---|---|---|
| 0 | 03.08.2011 03:08:30 | 112141 | New reservation | fab |
| 1 | 19.08.2011 16:09:35 | 112438 | New reservation | lov |
| 2 | 19.08.2011 16:09:35 | 112439 | New reservation | lov |
| 3 | 16.09.2011 15:48:36 | 112360 | New reservation | fab |
| 4 | 16.09.2011 15:48:43 | 112361 | New reservation | fab |
| 5 | 23.09.2011 12:47:59 | 112012 | New reservation | lov |
| 6 | 23.09.2011 12:47:59 | 112013 | New reservation | lov |
| 7 | 23.09.2011 12:48:04 | 112014 | New reservation | top |
| 8 | 23.09.2011 12:48:04 | 112015 | New reservation | top |
| 9 | 03.10.2011 15:43:17 | 112265 | New reservation | lov |

**Figure 2.** Sample of the events recorded in the actual event log.

**Figure 4.** Correlations between attributes available in the event log.

Event log data were collected and combined from many different database tables in the PMS relational database management system. In order to gain an insight into the log data, we performed an exploratory analysis proposed in [38]. Overview information about the event log used in this paper is shown in Figures 2–4. The raw event log data is a CSV file that contains 2124 events (Figure 2), described using four attributes, without any missing data: TASKID, TASKNAME (five distinct activities shown in Figure 3a), USERNAME (15 distinct users shown in Figure 3b) and DATETIME.

There are 437 TASKIDs that represent individual guests. Events with the same TASKID will be grouped in individual traces. It can be seen from Figure 4 that there is strong correlation between attributes TASKNAME and USERNAME, because some users always perform certain tasks, e.g. telephone exchange always adds extra services, back office makes reservations and invoices to agencies, and employees at the reception check in guests, perform billing and check outs.

Event data are extracted from the event log's CSV file using Python *pm4py* library[39] and Python[1] *pandas* data frame structure. In order to identify traces, mapping of attributes is performed by setting corresponding parameters. Special attention was paid to the timestamp format conversion. Events grouped in traces are stored as an extension of the *Python* list and saved into a text file. We have written a *Python* script that automates these data transformations. Additionally, we wrote *Perl* script that prepares traces for further processing by our proposed process synthesis method. Overview information about traces is shown in Figures 5 and 6. Due to reasons of clarity, events in traces have abbreviated names.

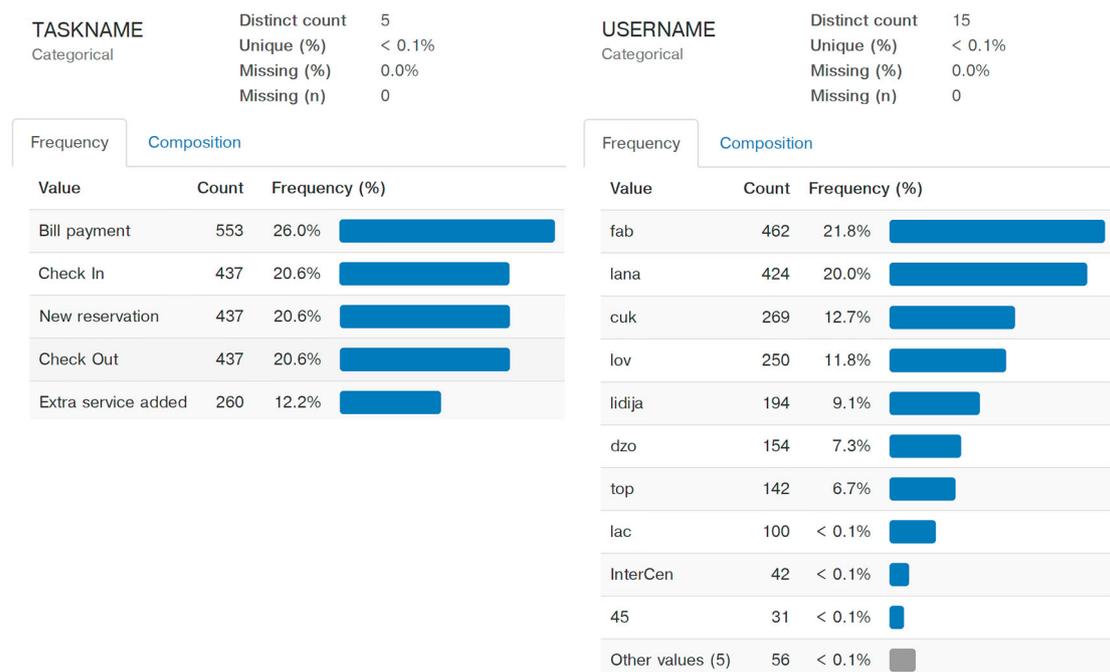

**Figure 3.** (a) Overview of the attribute TASKNAME and (b) overview of the attribute USERNAME.

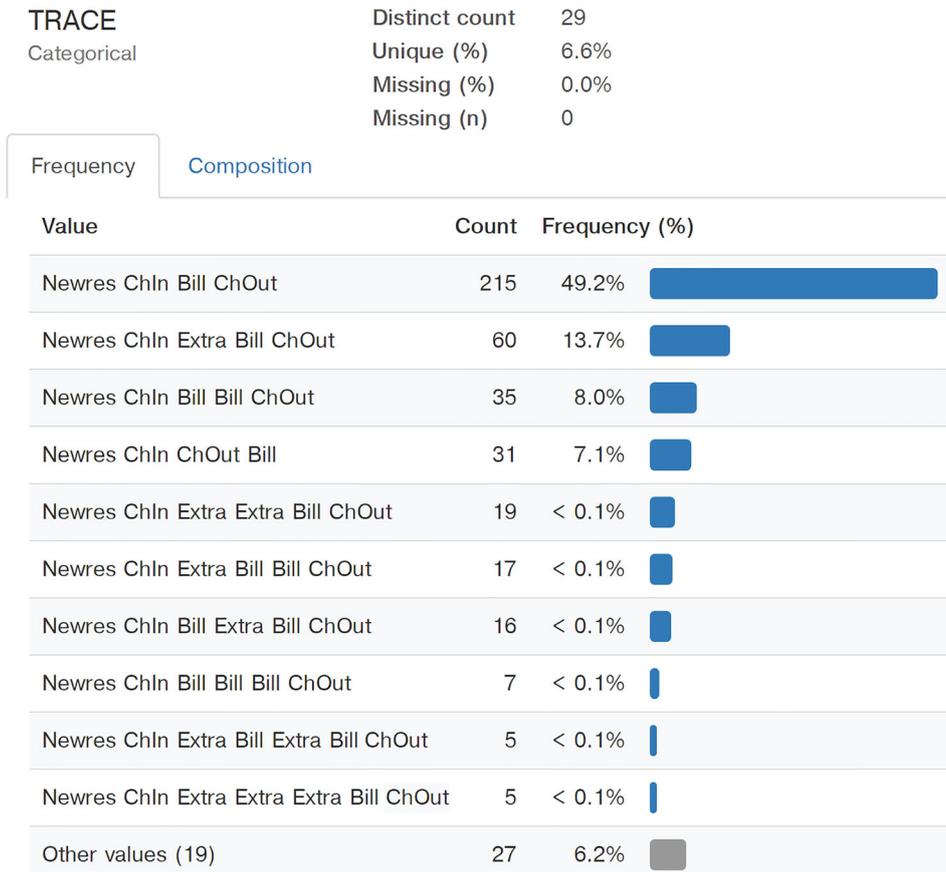

**Figure 5.** Overview of traces recorded in the event log data.

| | TASKID | TRACE |
|---|---|---|
| 0 | 112141 | Newres ChIn Extra Extra Bill Bill ChOut |
| 1 | 112438 | Newres ChIn Bill Bill ChOut |
| 2 | 112439 | Newres ChIn Bill Bill ChOut |
| 3 | 112360 | Newres ChIn Bill ChOut |
| 4 | 112361 | Newres ChIn Bill ChOut |
| 5 | 112012 | Newres ChIn Bill ChOut |

**Figure 6.** A sample of the traces from the event log data.

## 4. Automated analysis of discovered process models using model checking methods

The process models we discovered from the preprocessed log data using our process discovery method are finite state automatons, encoded as directed graphs with labelled nodes and edges in *Graphviz*[2] dot textual format. Therefore, they are static files that can be exported to standard image or document formats, e.g. JPEG, PNG, SVG or PDF. As an example, Figure 7 shows the initial imperfect process model and a more complex process model, both discovered by our process discovery method from the same input log data. Process mining experts can use these images to check and validate process models manually. Nevertheless, even for simpler process models (e.g. Figure 7a) we should not always rely only on manual and visual inspection methods.

Researchers have explored various ways for automated checking of process models, e.g. by transforming the process model finite state automaton to a Petri net [40,41] that can be dynamically simulated and checked for problems. In our paper, we will use model checking techniques to automatically simulate and verify the discovered process models.

### 4.1. Model checking

Model checking is a formal method for software and hardware system verification. Its goal is to check whether a model of a system satisfies given specification. We will use the Spin model checker that was developed at the Bell Labs by G.J. Holzman [42]. Spin model checker is primarily used for formal verification of distributed systems, such as communication protocols. Spin can run random simulations of the process model or perform a verification of the process model by exploring all the possible execution paths. To formally describe process models, we use Spin's *Promela* language (*Process meta language*). Some basic elements of the *Promela* syntax are similar to the standard C language, but with a different semantics that implicitly enables non-determinism and statement blocking in the execution of the program code.

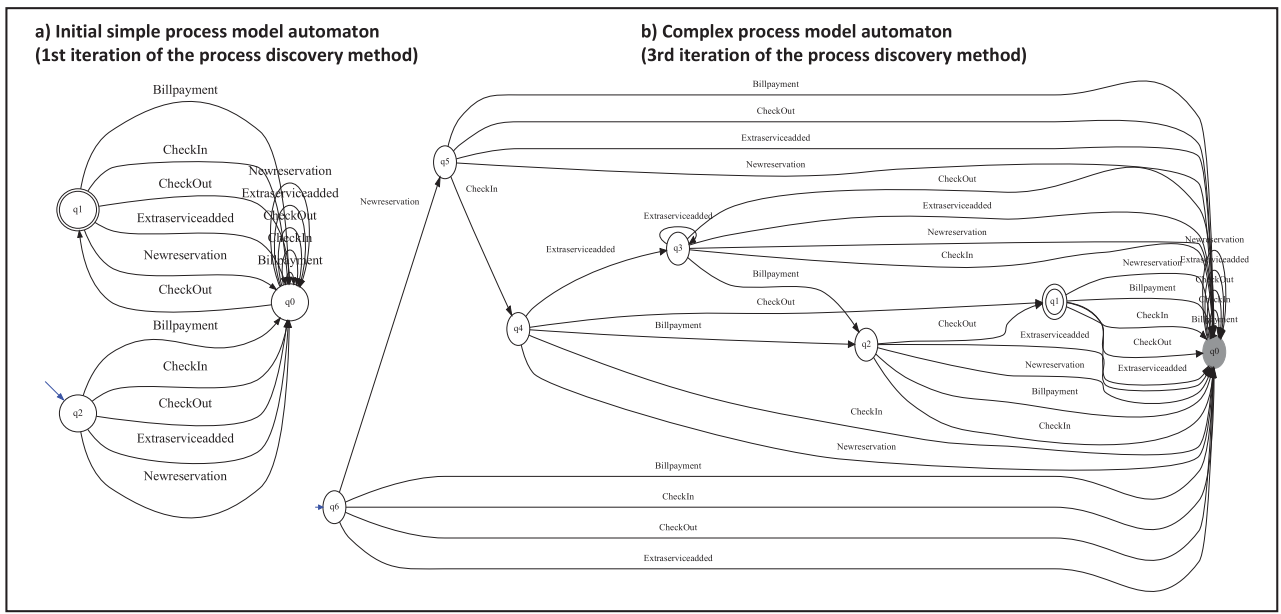

**Figure 7.** (a) Simple process model automaton (three states); (b) complex process model automaton (seven states).

For example, the order of execution in the conditional *if–then–else* statement is not predetermined – Spin will non-deterministically execute one of the branches that are not blocked. Specifications that are used to check model correctness are invariants – statements that must be true during the entire execution run. They can be defined as *assert* statements or more formally as Linear temporal logic (LTL) formulae [16,42]. LTL is a modal logic that uses temporal operators, e.g. finally and globally in addition to common logical operators, e.g. negation, conjunction, disjunction and logical implication.

### 4.2. Automated model transformation into Promela language

But, to use this model checking method we first need to transform the discovered process models in dot format into the *Promela* language. For simple process models, this can be done by hand, although it can be cumbersome and prone to errors when transforming larger process models. Therefore, we have devised an algorithm implemented as a *Perl* script that automates this transformation procedure. Proposed model transformation algorithm from dot format to *Promela* file is presented in Figure 8.

Our script parses through the initial process model in the dot format and using a set of regular expressions extracts all the necessary data for the construction of the *Promela* process model. Dot format stores information about the initial state (node denoted with a blue ingoing arrow), regular end state (node denoted as a double circle), sink state (i.e. irregular end state, node denoted as a grey-filled circle) and other states (rest of the nodes denoted as circles), as well as all the labelled transitions between the states. Our script groups all the transitions with the same starting state in a single *if-then-else* block and transforms all the transition labels to *printf* statements. Script also ensures that the first *if-then-else* block in the *Promela* process model contains transitions from the initial state. Additionally, script inserts a global variable *end_state* that is initialized to 0 and set to 1 only when the end state is reached. To improve readability of simulation results script inserts control *printf* statement each time the end state or the sink state is entered. Also, script automatically adds an additional *if-then-else* option that allows simulation execution to stop when it enters the end state or the sink state. This modification of the process model ensures simulation runs will not be infinite. Lastly, script adds a *never* claim that implements a simple LTL formula that checks if the model reaches regular end state.

It must be noted that this automatically created *Promela* model can be simulated and verified using Spin model checker without further modifications.

### 4.3. Simulation of the Promela process model

Spin model checker supports three main simulation types: the default random simulations, interactive simulations and the guided simulation. In a random simulation, Spin automatically traverses through the process model(s), and that path is chosen non-deterministically. The interactive simulation requires that the user always chooses the next step from a list of the available options. This can be useful for process mining experts when exploring and checking some interesting or infrequent paths through the process model automaton. Guided simulation shows the path to the error encountered during the verification run. If the verification fails an error *trail* file is generated

```
Algorithm dot_to_Promela:
initialize empty promela_model
open(dot_file)
read_line_by_line(dot_file) {
        detect_end_state
        detect_sink_state
        detect_initial_state
        detect_transitions(from_state, label, to_state) {push to transitions[from_state]: print_label_goto_to_state}
}
close(dot_file);
add_default_headers(promela_model)
add_if_block(promela_model, initial_state) {
        promela_model += if_block_header
        foreach element in transitions[initial_state] { promela_model += element }
        promela_model += if_block_footer
}
for i=0 to transitions.number
        if i != Initial_state {
                add_if_block(promela_model, i)
                        promela_model += if_block_header
                        if i == end_state { promela_model += print_END }
                        else if i == sink_state { promela_model += print_SINK }
                        foreach element in transitions[i] { promela_model += element }
                        if i == end_state || i == sink_state { promela_model += goto_stop_state }
                        promela_model += if_block_footer
                }
        }
}
add_stop_state(promela_model)
add_deafult_footer(promela_model)
add_default_never_block_LTL_formula(promela_model, "finally p")
save(promela_model)
```

**Figure 8.** Algorithm for process model transformation from dot file into Promela file.

that records all the steps taken on the path to the error. To replay and examine that trail file, process mining experts can use Spin runtime option *-t*.

### 4.4. Automatic generation of positive and negative examples

The simulation results suggest that the Spin model checker can be used as a random generator of both positive and negative examples. Nevertheless, random generator will only work effectively if the underlying process model is adequately selective. Therefore, process mining experts should start with a more complex and representative process model, e.g. a finite state automaton discovered in the third iteration of our process model discovery method (Figure 7b). In the first step of the classification process, each trace that enters the end state can be automatically classified as a positive example, and all others can be classified as negative examples.[3] Afterward, process mining experts could also manually inspect these classification results to possibly detect false positive or false negative examples. All the gathered information about the true/false positive and true/false negative examples could be then used in a feed-back loop as an additional input to our process discovery method that can help to improve or repair the discovered process model finite state automatons.

```
#!/bin/sh

I=promela_model_file
rm -f $I-gen.traces
echo "WORKING ...$I-gen.traces "
j=$RANDOM
for i in {1..1000}
do
        spin -n$j $I.pml >> $I-gen.traces
        j=$((j + 1))
done
sed -i 's/1 process created//g' $I-gen.traces
sed -i 's/          / /g' $I-gen.traces
sed -i 's/^[ \t]*//g' $I-gen.traces
sed -i '/warning:/d' $I-gen.traces
echo "DONE"
```

**Figure 9.** Shell script traces-generator.sh runs and stores results of 1000 spin simulations.

We have written Linux *shell* scripts that help to automate these processes. The first *shell* script *traces-generator.sh* (Figure 9) facilitates the running and storing results of a requested number of simulations. The second *shell* script *positive-negative-traces-generator.sh* (Figure 10) generates a list of positive examples (when the traces reach end state) or negative examples (when the traces do not reach end state). It runs the simulations in an infinite loop and stops when the requested number of positive or negative examples has been reached.

It must be noted that precautions must be made when running Spin simulations inside a loop. By

```
#!/bin/sh

I=promela_model_file
echo "WORKING ...$I-gen.traces "
examples_type="positive"
j=$RANDOM
k=0
number=5
result=0
found=0
simulation=""
rm -f $I-$examples_type.traces
while true
do
        simulation=$(spin -n$j $I.pml)
        if [[ $simulation =~ "END" ]]; then found=1
        else found=0
        fi
        if [[ ${examples_type} == "positive" && ${found} -gt 0 ]] || \
           [[ ${examples_type} == "negative" && ${found} -eq 0 ]]
        then
                echo "$simulation" >> "$I-$examples_type.traces"
                k=$((k + 1))
                if [ $k == $number ]; then break
                fi
        fi
        j=$((j + 1))
done
sed -i 's/1 process created//g' $I-$examples_type.traces
sed -i 's/       / /g' $I-$examples_type.traces
sed -i 's/^[ \t]*//g' $I-$examples_type.traces
sed -i '/warning:/d' $I-$examples_type.traces
echo "DONE"
```

**Figure 10.** Shell script positive-negative-traces-generator.sh generates and stores five positive examples.

default, Spin uses a seed for the random number generator based on the system timestamp. In our case, the loop iterated very quickly, and therefore the seed for successive Spin simulations did not change which resulted in the same simulation results. To fix this unwanted behaviour, we used a Spin option -nN that explicitly sets different seed number *N* in every iteration of the loop.

### 4.5. Promela process model verification

Even by running numerous simulations of the *Promela* process model, there are no guarantees that an error in the model will be detected. However, Spin model checker does not only support simulation of the *Promela* model but also allows for its rigorous verification by checking if the model satisfies given specifications. First, Spin model checker automatically generates a process analyser C program code (*pan.c*) for the given *Promela* process model. Process analyser (also called a *verifier*) is then compiled using standard C compilers (e.g. *gcc*). The verifier performs automatic verification of the *Promela* model by examining all possible execution paths. To reduce memory requirements, Spin uses special bitstate hashing functions. It must be noted when no specification is given *pan* verifier still checks if there is a deadlock in the *Promela* process model.

To define simple specifications, we can use the *assert* statement that takes an argument that must be evaluated as *true*. This argument can be considered as an *invariant* of the system model, whose value should not change during the verification process. Though, if that value changes then the assert statement will be violated, and the verification process fails.

When it is necessary to check more complex specifications, LTL formulae can be used. Spin automatically generates a *never* claim, special process type that implements given LTL formula as a Büchi automaton. If the never claim process completes (reaches end state or enters an $\omega$-acceptance infinite cycle), then the given LTL formula is satisfied. At that moment, verification stops with an error and verifier generates an error trail, also called *a counterexample*. Spin implicitly determines the synchronous product of the defined process model(s) and the never claim, which guarantees that the never claim will be evaluated at each step of the process model execution.

In our model transformation algorithm (Figure 8), we specified default LTL formula $\Diamond p$ (*finally p*), where $p$ is a predicate $p \equiv end\_state == 1$. This LTL formula checks if the process model will eventually reach the regular end state. Figure 11 shows the simplified Büchi automaton of this LTL formula and its never claim in *Promela*, generated automatically using Spin LTL conversion option -f "$< > p$".

Figure 11(a) shows that the LTL formula $\Diamond p$ is satisfied (i.e. the Büchi automaton end state $S_1$ is reached) if and only if predicate $p$ becomes true somewhere along the process model execution paths.

Since the version Spin 6.0.0 LTL formulae can be also specified directly inline using new *Promela* command *ltl*. It must be noted that inline LTL formulae always describe positive properties. Therefore, any specified inline LTL formula will be automatically negated so a counterexample could be found during the verification run.

However, in this paper we have used older mechanism of *Promela* never claims, because we considered that all our LTL formulae describe negative properties. Due to this, our LTL formulae are not negated before conversion to the *Promela* never claims. This approach also ensures that the verification run will generate a counterexample when such LTL formula is satisfied.

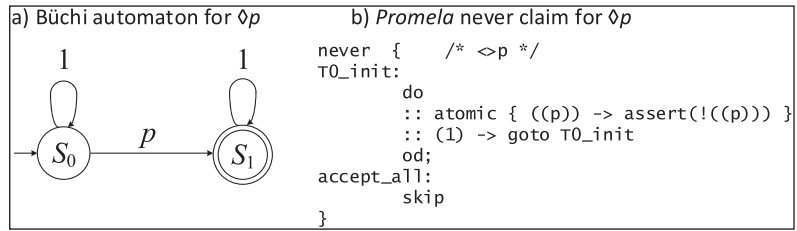

**Figure 11.** (a) Büchi automaton for LTL formula < > p; (b) corresponding Promela never claim.

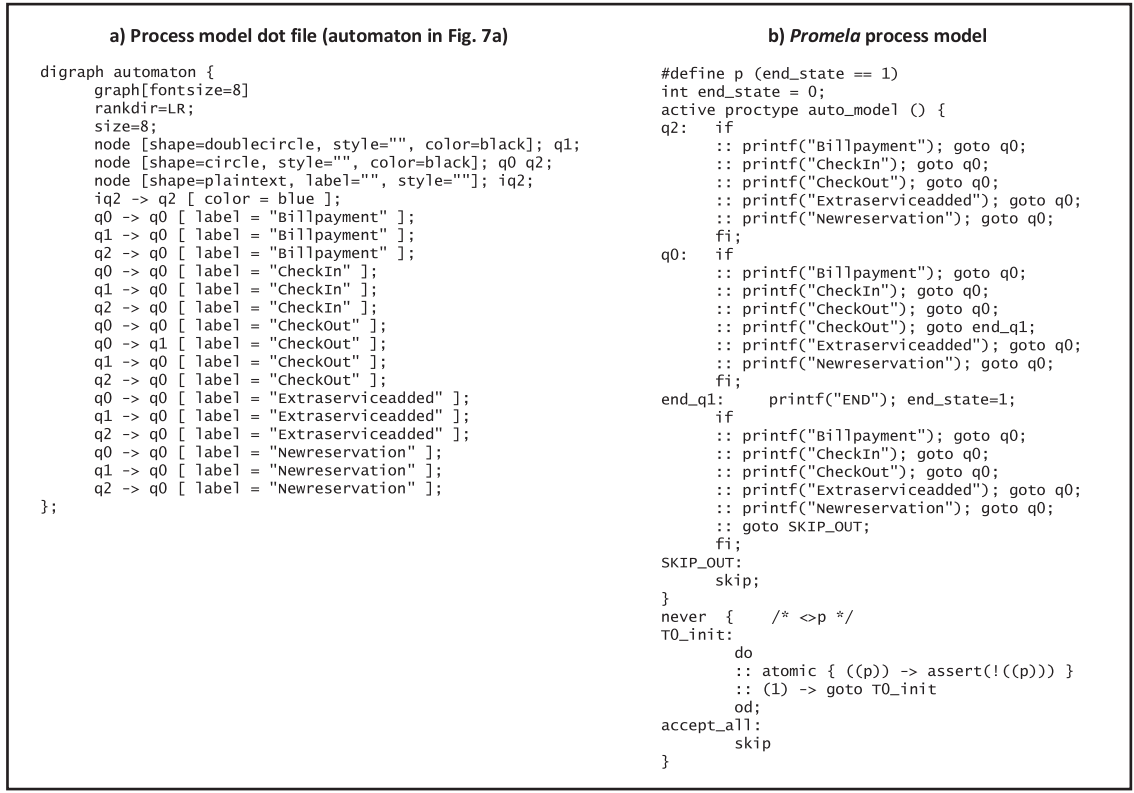

**Figure 12.** (a) Initial process model from Figure 7(a) in dot format; (b) corresponding automatically generated Promela code.

## 5. Results and discussion

In this section, we will present and discuss results of the analysis of process models discovered from event log data using model checking methods. We will analyse two discovered process models shown in Figure 7 – simpler initial process model (Figure 7a) and more complex and representative process model (Figure 7b). The simulations and verifications were run on a desktop computer (Intel i3 CPU 540, 8 GB RAM) inside *Cygwin* environment on a Windows OS. Linux system command *time* was used to measure performance. We used Spin version 6.5.0 for *Promela* process model simulations and verifications.

### 5.1. Promela process model simulation results

As an example of our proposed model transformation algorithm into *Promela* language (Figure 8), we show in Figure 12 side by side comparison of the simple process model in dot format from Figure 7(a) and the corresponding automatically generated *Promela* process model.

By examining the structure of the *Promela* process model generated by our algorithm, it can be observed that every labelled transition can be written to standard output during the simulation execution in Spin. Therefore, the output of each simulation run is a list of executed *printf* commands. That list is a trace (possible sequence of events) or, stated more formally – one random path through the discovered process model finite state automaton (e.g. Figure 7a).

Figure 13 shows the results of simulations of the *Promela* process model in Figure 12(b) using default Spin settings. These simulation results corroborate our earlier claim that this initial process model is not

Irregular traces (results of the random simulations):
**1st** *CheckOut     Extraserviceadded     Billpayment     Newreservation     Billpayment     CheckIn     CheckIn     Newreservation     CheckIn     Newreservation     Extraserviceadded     CheckOut     END*

**2nd** *CheckOut     Billpayment     Billpayment     CheckOut     END*

**3rd** *Extraserviceadded     Newreservation     Extraserviceadded     Billpayment     Extraserviceadded     CheckOut     END*

**4th** *CheckIn     Newreservation     CheckOut     CheckOut     CheckIn     Newreservation     CheckIn     CheckIn     CheckOut     END*

Regular trace (result of the interactive simulation):
**5th** *Newreservation     CheckIn     Billpayment     CheckOut     END*

**Figure 13.** Selection of five distinct simulation runs of the Promela process model in Figure 12(b).

**Table 1.** Cumulative results of 10 executions of the simulation script (requested number of traces from 100 to 1000).

| | | | | | | | | | | |
|---|---|---|---|---|---|---|---|---|---|---|
| Generated number of traces | 100 | 200 | 300 | 400 | 500 | 600 | 700 | 800 | 900 | 1000 |
| Script runtime (s) | 10.93 | 21.12 | 31.55 | 42.33 | 54.31 | 63.81 | 73.53 | 84.92 | 94.46 | 108.1 |
| Percentage of CPU used | 9 | 8 | 9 | 8 | 8 | 8 | 8 | 8 | 8 | 8 |
| Trace file size (kB) | 10.2 | 21.9 | 30.3 | 43.1 | 52.1 | 62.6 | 71.9 | 84.7 | 95.4 | 109 |
| Number of positive examples | 0 | 2 | 3 | 5 | 6 | 6 | 5 | 7 | 9 | 10 |
| Number of negative examples | 100 | 198 | 297 | 395 | 494 | 594 | 695 | 793 | 891 | 990 |
| Ratio positive/negative examples (%) | 0 | 1.01 | 1.01 | 1.27 | 1.21 | 1.01 | 0.72 | 0.88 | 1.01 | 1.01 |
| Shortest trace length | 1 | 1 | 1 | 1 | 1 | 1 | 1 | 1 | 1 | 1 |
| Longest trace length | 32 | 29 | 28 | 48 | 48 | 38 | 41 | 34 | 35 | 42 |
| Average trace length | 6.01 | 6.47 | 5.96 | 6.38 | 6.15 | 6.17 | 6.07 | 6.26 | 6.26 | 6.42 |

perfect, because both regular traces (fifth simulation run) and irregular traces (simulation runs 1–4) reach end state. Therefore, this process model should not be used as a basis for the generator of positive and negative examples because it has a high rate of false-positive examples (irregular traces that reach end state). It is also evident from Figure 13 that the simulation results are not uniform – the length of the traces varies from 4 to 12 (not counting the control END *printf* statements).

Afterward, the process model in Figure 7(b) was also automatically transformed from the *dot* format to the *Promela* language. This process model is more complex than the one in Figure 7(a), and also note that it has a sink state. Therefore, it is suitable to use as the basis for generating positive and negative examples using our automated simulation *shell* script *traces-generator.sh* in Figure 9. Table 1 contains data on cumulative results of 10 executions of this simulation script.

In first call of the simulation script *traces-generator.sh*, we have set the number requested traces to be 100, and in subsequent 9 calls we increased this number iteratively by 100. It can be observed from the presented data that the script runtime increased almost linearly from 10.93 s to 108.1 s as we increased the number of generated traces from 100 to 1000. We can see that the shortest trace length (excluding control *printf* statements) in all 10 simulation script runs was only 1, and that the longest trace length varied from 28 to 48. Vast majority of the generated traces in all 10 simulation script runs were classified as negative examples.

This can be explained by examining the underlying process model automaton in Figure 7(b). There are numerous paths that can be taken from the initial state $q_6$ to the grey-filled sink state $q_0$ (irregular end state), and only few of the possible paths lead to the regular end state $q_1$. Indeed, in the first simulation script run there were no positive examples generated at all. In the subsequent simulation script runs, there was approximately 1% positive examples among all generated traces. Overall, there were 53 positive examples generated during the 10 runs of the simulation script, but they can all be reduced to three unique traces in Table 2.

For demonstration of the second *shell* script *positive-negative-traces-generator.sh* (Figure 10), we also used automatically generated *Promela* process model of the discovered automaton in Figure 7(b). First, the script was set to generate 300 negative examples. During the execution time of 33.6 s, the script ran 304 Spin simulations and filtered traces for negative examples. Among the 300 generated traces, there were 243 unique negative examples and 57 duplicates. Afterward, the script was set to generate 10 positive examples. This time, the execution time was longer (79.63 s), because the script needed to run 713 Spin simulations to find 10 positive examples. Between them, there were three unique positive examples: {*Newreservation*, *CheckIn*, *CheckOut*}, {*Newreservation*, *CheckIn*, *Billpayment*, *CheckOut*} and {*Newreservation*, *CheckIn*, *Extraserviceadded*, *Extraserviceadded*, *Billpayment*, *CheckOut*}.

**Table 2.** Overview of positive examples generated after 10 executions of the simulation script.

| Positive example trace | | | | | Frequency |
|---|---|---|---|---|---|
| Newreservation | CheckIn | CheckOut | | | 39 |
| Newreservation | CheckIn | Billpayment | CheckOut | | 13 |
| Newreservation | CheckIn | Extraserviceadded | Billpayment | CheckOut | 1 |

**Table 3.** Results of six verification runs for Promela process model of the automaton in Figure 7(b).

| Run | Correctness specification | Predicates | Detected counterexample | Performance measurements |
|---|---|---|---|---|
| 1st | assert(p) | $p \equiv$ end_state $== 0$ | Newreservation, CheckIn, CheckOut | 0.082 s<br>64.54 MB |
| 2nd | LTL formula $\diamond p$<br>(finally p) | $p \equiv$ end_state $== 1$ | Newreservation, CheckIn, Billpayment, CheckOut | 0.078 s<br>64.54 MB |
| 3rd | LTL formula $\diamond p$<br>(finally p) | $p \equiv$ end_state $== 1$ &&<br>extra_service $== 2$ | Newreservation, CheckIn, Extraserviceadded, Extraserviceadded, Billpayment, CheckOut | 0.142 s<br>65.71 MB |
| 4th | LTL formula $p \rightarrow \diamond q$<br>(if p then finally q) | $p \equiv$ checkIn $== 1$ \|\|<br>end_state $== 0$<br>$q \equiv$ checkOut $== 1$ &&<br>end_state $== 1$ | Newreservation, CheckIn, CheckOut | 12.263 s<br>435.144 MB |
| 5th | LTL formula $p$ U $q$<br>(p holds at least until q holds) | $p \equiv$ end_state $== 0$<br>$q \equiv$ end_state $== 1$ &&<br>bill_payment $== 1$ | Newreservation, CheckIn, Billpayment, CheckOut | 0.078 s<br>64.83 MB |
| 6th | LTL formula $\diamond p$(finally p) | $p \equiv$ end_state $== 1$ &&<br>sink_state $== 0$ &&<br>bill_payment $== 3$ | No counterexample found! | 22.136 s<br>1008.397 MB |

## 5.2. Promela process model verification results

Verification was performed on the same complex *Promela* process model that was used for automated generation of positive and negative examples (Figure 7b). We defined six different specifications, five using LTL formulae and one using *assert* statement. As stated earlier, we used older mechanism of *Promela* never claims for specifying LTL formulae. Also, we considered that all five LTL formulae describe negative properties and therefore they are not negated before conversion to the *Promela* never claims. Table 3 contains the data of the verification runs, each with the used correctness specification, a counterexample if it was detected, as well as performance measurements (verification run-time and amount of RAM used).

In the first verification run in Table 3, we wanted to know if the model can reach the regular end state. Therefore, we supposed the opposite in an assert statement. Verification failed and provided one counterexample where the model indeed enters the regular end state. In the subsequent verification runs from Table 3 only LTL formulae were used. Verification will fail and provide a counterexample if and only if the specified LTL formula is satisfied. In the second verification run, we again check that the end state can be reached, but the verifier generated a longer counterexample. Before running remaining verifications slight modifications of the *Promela* model have been performed by adding global variables that counted certain events (number of check-ins, check-outs, bill payments and added extra services) or registered reaching sink state. It must be noted that these modifications increase the complexity of the model and potentially expand the search space the verifier must explore. This is especially the case in the fourth verification run, where the LTL formula p $\rightarrow \diamond q$ is equivalent to $\neg p \vee \diamond q$. Therefore, verifier must find all states where $\neg p$ holds (reached regular end state without a single check-in event) or finally $q$ holds (maximum one check-out event before regular end state is reached). The LTL formula was satisfied, and the verifier returned a counterexample where, as expected, both check-in and check-out events occurred once before regular end state was reached. In the last, sixth verification run we check is it possible to count three bill payments events before entering regular end state. Verifier could not satisfy that LTL formula, even after the maximum search depth was increased from the default $10^4$ steps to $10^8$ steps. Therefore, in this case no counterexample was generated. This conclusion could be used by process mining expert for repair of the discovered process model in Figure 7(b), so that it accepts multiple bill payment events before reaching the regular end state.

Finally, we have demonstrated that using Spin verification mode process mining experts can check the existence or absence of various specific traces, both regular and irregular. Therefore, those traces should be injected in the training and the test data marked as positive, or negative examples, respectively. Then, process repair can be performed by repeating the process discovery and analysis until the optimal model is obtained in the context of the balanced quality metrics (*fitness*, *precision*, *generalization*, *simplicity*) [4]. Also, this could be done by using other techniques to add/remove those paths in process model, e.g. process repair based on counterexamples, which is the plan of our future work.

## 6. Conclusion

In this paper, we gave an overview of our proposed method for process synthesis, analysis and repair. We presented in detail the results of our model checking based method for process analysis. As a use-case example, we used actual data gathered from a hotel data management system. It was demonstrated that Spin model checker can be successfully used to automatically simulate and verify discovered process models. Verification was performed under six different specifications, five using LTL formulae and one using *assert* statement. The obtained results confirmed the efficiency of the proposed approach. In addition to verification of process model soundness, using our approach experts can detect irregular traces (false positive examples), generate negative examples, check if the sequence of events is satisfied in process model executions, as well as other specifications.

The advantage of our approach is that it provides automated analysis of the process mining results by simulation and verification of process model executions. Process analysis results will be used in further research to improve and refine discovered process models. Repairing process models is necessary due to consistency issues or behavioural problems, and our approach provides foundations for both types of repair.

The presented method can be used for the detection and analysis of expected and unexpected patterns in the process and system behaviour of IoT devices that constantly generate streams of event data.

## Acknowledgements

We would like to thank Romualdo Miljak who prepared the event log CSV file used in this paper and provided domain expertise. This research was supported by the University of Dubrovnik VIF funds. The authors would like to thank the reviewers for their precious time and valuable insights.

## Disclosure statement

No potential conflict of interest was reported by the author(s).

## Notes

1. https://www.python.org/
2. https://www.graphviz.org/
3. *Under such classification, for models without a sink state (e.g., for the automaton in* Figure 7a), *all simulation runs are positive examples because a simulation run can terminate only in an end state.*

## References

[1] van Zelst SJ, van Dongen BF, van der Aalst WM. Event stream-based process discovery using abstract representations. Knowl Inf Syst. 2018;54(2):407–435.

[2] Sagiroglu S, Sinanc D. Big data: a review. 2013 International Conference on Collaboration Technologies and Systems (CTS); 2013. p. 42–47.

[3] Dhar V. Data science and prediction. Commun ACM. 2013;56(12):64–73.

[4] van der Aalst WMP. Data science in action. New York, NY: Springer; 2016.

[5] van der Aalst WMP. Viewing the internet of events through a process lens. In: BPM everywhere. Future Strategies; 2015. p. 213–221.

[6] Zakarija I, Škopljanac-Macina F, Blaškovic B. Discovering process model from incomplete log using process mining. In: 2015 57th International Symposium ELMAR (ELMAR); İEEE; 2015. p. 117–120.

[7] Garcia CdS, Meincheim A, Faria Junior ER, et al. Process mining techniques and applications – a systematic mapping study. Expert Syst Appl. 2019;133:260–295.

[8] Zhang X, Du Y, Qi L, et al. An approach for repairing process models based on logic petri nets. IEEE Access. 2018;6:29926–29939.

[9] Xu J, Liu J. A profile clustering based event logs repairing approach for process mining. IEEE Access. 2019;7:17872–17881.

[10] Augusto A, Conforti R, Dumas M, et al. Automated discovery of process models from event logs: review and benchmark. CoRR. 2017;6:24543–24567. abs/1705.02288.

[11] Erdogan TG, Tarhan A. Systematic mapping of process mining studies in healthcare. IEEE Access. 2018;6.

[12] De Weerdt J, De Backer M, Vanthienen J, et al. A multi-dimensional quality assessment of state-of-the-art process discovery algorithms using real-life event logs. Inf Syst. 2012;37(7):654–676.

[13] Bolt A, de Leoni M, van der Aalst WM. Scientific workflows for process mining: building blocks, scenarios, and implementation. Int J Software Tools Technol Transfer. 2016;18(6):607–628.

[14] Elgammal A, Turetken O, van den Heuvel W-J, et al. Formalizing and applying compliance patterns for business process compliance. Software Syst Model. 2016;15(1):119–146.

[15] Chesani F, Lamma E, Mello P, et al. Exploiting inductive logic programming techniques for declarative process mining. In: Transactions on Petri Nets and Other Models of Concurrency II; Springer; 2009. p. 278–295.

[16] van der Aalst WMP, Beer HT, Dongen BF. Process mining and verification of properties: An approach based on temporal logic. In: On the Move to Meaningful Internet Systems 2005: CoopIS, DOA, and ODBASE, LNCS; Springer; 2005. vol. 3760, p. 130–147.

[17] Lamma E, Mello P, Riguzzi F, et al. Applying inductive logic programming to process mining. In: International Conference on Inductive Logic Programming; Springer; 2007. p. 132–146.

[18] Räim M, Di Ciccio C, Maggi FM, et al. Log-based understanding of business processes through temporal logic query checking. In: OTM Confederated International Conferences "On the Move to Meaningful Internet Systems"; Springer; 2014. p. 75–92.

[19] Janssen M, Mateescu R, Mauw S, et al. Verifying business processes using Spin. In: Proceedings of the 4th International SPIN Workshop; 1998. p. 21–36.

[20] van der Aalst WM, Adriansyah A, de Medeiros AKA, et al. Process mining manifesto. In: Business Process Management Workshops; Springer; 2012. p. 169–194.

[21] R'bigui H, Cho C. The state-of-the-art of business process mining challenges. Int J Bus Process Integr Manage. 2017;8:285.

[22] "IEEE CIS Task Force on Process Mining."[cited 2019 Nov 23]. Available from: https://www.win.tue.nl/ieeetfpm/doku.php?id = start.

[23] Dumas M, Fantinato M, eds. Business Process Management Workshops: BPM 2016 International Workshops, Rio de Janeiro, Brazil, September, 2016, LNBIP. Springer, 2017.

[24] Business Process Management Workshops BPM 2017. New York, NY: Springer, 2018.

[25] 2019 International Conference on Process Mining ICPM 2019. Aachen, Germany: IEEE, June 2019.

[26] van der Aalst W. Keynote: 20 years of process mining research accomplishments, challenges, and open problems. In: 2019 International Conference on Process Mining (ICPM); June 2019. p. 14–14.

[27] Bollig B, Katoen J-P, Kern C, et al. Libalf: The automata learning framework. In: Computer Aided verification; LNCS. Springer; 2010. vol. 6174, p. 360–364.

[28] D'Souza D, Shankar P, eds. *Modern applications of automata theory*, ser. IISc research monographs series, Vol. 2. Princeton (NJ): World Scientific; 2012.

[29] Biermann AW, Feldman JA. On the synthesis of finite-state machines from samples of their behavior. IEEE Trans Comput. 1972;C-21(6):592–597.

[30] van der Aalst WMP. Discovering coordination patterns using process mining. In: Workshop on Petri Nets and Coordination; Bologna, Italy; 2004. p. 49–64.

[31] Li W, Zhu H, Liu W, et al. An anti-noise process mining algorithm based on minimum spanning tree clustering. IEEE Access. 2018;6:48756–48764.

[32] Domingos P. A few useful things to know about machine learning. Commun ACM. 2012;55(10):78–87.

[33] Dziak D, Jachimczyk B, Kulesza W. IoT-based information system for healthcare application: design methodology approach. Applied Sci. 2017 Jun;7:596.

[34] Minbo L, Zhu Z, Guangyu C. Information service system of agriculture IoT. Automatika. 2013;54(4):415–426.

[35] Crnkovic I, Sentilles S, Vulgarakis A, et al. A classification framework for software component models. IEEE Trans Software Eng. 2011;37(5):593–615.

[36] Blaškovic B, Ježic G, eds. *Special issue on agent and multiagent systems design*, ser. CIT J Comput Inf Technol. 2019;27(1).

[37] Burattin A. Applicability of process mining techniques in business environments [Ph.D. dissertation]. University of Bologna; 2013.

[38] Humski L, Pintar D, Vranic M. Exploratory analysis of pairwise interactions in online social networks. Automatika. 2017;58(4):422–428.

[39] Berti A, van Zelst SJ, van der Aalst W. Process mining for Python (PM4Py): Bridging the gap between process-and data science. CoRR. 2019:13–16. abs/1905.06169

[40] de Medeiros AKA, van der Aalst W, Weijters A. Workflow mining: current status and future directions. In: On the move to meaningful internet systems 2003: CoopIS, DOA, and ODBASE, volume 2888 of LNCS; Springer-Verlag; 2003. p. 389–406.

[41] van der Aalst WM, Weijters T, Maruster L. Workflow mining: Discovering process models from event logs. IEEE Trans Knowl Data Eng. 2004;16(9):1128–1142.

[42] Holzmann GJ. The SPIN model checker: primer and reference manual. Reading: Addison-Wesley; 2004; vol. 1003.